\documentclass[conference, letterpaper]{IEEEtran}

\IEEEoverridecommandlockouts
\usepackage{cite}
\usepackage{amsmath,amssymb,amsfonts}
\usepackage{algorithmic}
\usepackage{graphicx}
\usepackage{textcomp}
\usepackage{tabularx}
\usepackage{booktabs}
\usepackage{multirow}
\usepackage{subcaption} 
\usepackage{float}      

\usepackage{marvosym}
\usepackage{hyperref} 

\usepackage{geometry}
\usepackage[table]{xcolor} 
\usepackage{colortbl} 

\def\BibTeX{{\rm B\kern-.05em{\sc i\kern-.025em b}\kern-.08em
    T\kern-.1667em\lower.7ex\hbox{E}\kern-.125emX}}
\begin{document}

\title{Towards Safe and Reliable Autonomous Driving: Dynamic Occupancy Set Prediction\\
}

\author{
Wenbo Shao, Jiahui Xu, Wenhao Yu, Jun Li, Hong Wang\textsuperscript{\Letter},~\IEEEmembership{Senior Member,~IEEE}
        \thanks{This work was supported by the National National Science Foundation of China Project: 52072215, 52221005, U1964203, and 52242213, National key R\&D Program of China: 2022YFB2503003, and State Key Laboratory of Intelligent Green Vehicle and Mobility.}
        \thanks{Wenbo Shao, Wenhao Yu, Jun Li and Hong Wang are with the School of Vehicle and Mobility, Tsinghua University, Beijing 100084, China. (e-mail: swb19@mails.tsinghua.edu.cn, wenhaoyu@mail.tsinghua.edu.cn, lijun1958@tsinghua.edu.cn, hong\_wang@mail.tsinghua.edu.cn)}
        \thanks{Jiahui Xu is with the School of Mechanical Engineering, Beijing Institute of Technology, Beijing 100081, China. (e-mail: 13645450063@163.com).}
        
}




\maketitle

\begin{abstract}
In the rapidly evolving field of autonomous driving, reliable prediction is pivotal for vehicular safety. However, trajectory predictions often deviate from actual paths, particularly in complex and challenging environments, leading to significant errors. To address this issue, our study introduces a novel method for Dynamic Occupancy Set (DOS) prediction, it effectively combines advanced trajectory prediction networks with a DOS prediction module, overcoming the shortcomings of existing models. It provides a comprehensive and adaptable framework for predicting the potential occupancy sets of traffic participants. 
The innovative contributions of this study include the development of a novel DOS prediction model specifically tailored for navigating complex scenarios, the introduction of precise DOS mathematical representations, and the formulation of optimized loss functions that collectively advance the safety and efficiency of autonomous systems.  Through rigorous validation, our method demonstrates marked improvements over traditional models, establishing a new benchmark for safety and operational efficiency in intelligent transportation systems.

\end{abstract}


\section{Introduction}

The rapid progression of autonomous driving technology is heralding a new era in transportation systems, bringing forth notable enhancements in traffic efficiency and vehicular safety. Yet, ensuring the safety and reliability of autonomous vehicles in the multifaceted real-world scenarios presents a significant challenge. This complexity arises not only from the advanced nature of the technology but also from the necessity to sustain system reliability in intricate environments.
The realm of trajectory prediction networks has undergone substantial development, becoming fundamental in boosting the predictive capabilities essential for autonomous vehicle navigation. These networks employ advanced algorithms and extensive datasets, enabling predictions of the movements of traffic participants by considering various factors like vehicle dynamics, road conditions, and traffic patterns.  Enhanced prediction accuracy and adaptability of these networks are particularly valuable in complex urban environments or situations involving unpredictable human interactions.   However, the complexity of real-world conditions and inherent limitations of predictive models \cite{wang2024survey} often lead to discrepancies between predicted and actual trajectories, posing a detriment to the reliable assurance of autonomous driving safety.

\begin{figure}
    \centering
    \includegraphics[width=8cm]{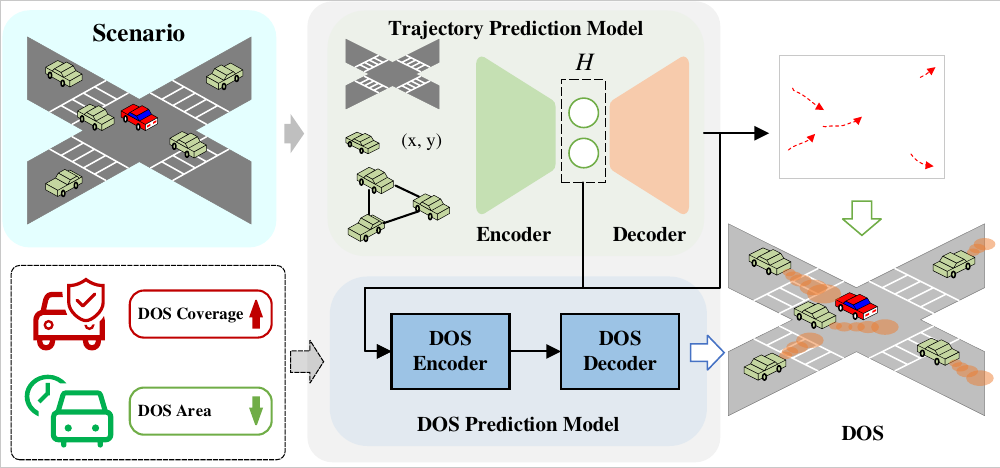}
    \caption{DOS prediction and its design objectives.}
    \label{fig1}
\end{figure}

In response to above challenges, our study introduces an novel Dynamic Occupancy Set (DOS) representation and estimation method for traffic participants.  DOS, which defines all potential positions a vehicle may occupy within a specified timeframe, plays a pivotal role in reinforcing the safety framework of autonomous driving systems. Traditional set-based prediction methods, often reliant on simplistic dynamic models or conservative assumptions, fall short in capturing the complexities of real-world driving scenarios. Our methodology, which integrates real-time environmental data with advanced trajectory prediction networks, enhances DOS predictions by minimizing the DOS area while maximizing coverage of traffic participants' future locations, thus balancing efficiency and safety, as shown in Fig. \ref{fig1}. Capable of adapting to evolving road conditions and traffic dynamics, this approach provides a comprehensive and nuanced risk assessment, which is essential for ensuring the safety of autonomous driving systems.

The principal contributions of this research are manifold:
\begin{itemize}
    \item \textbf{Integration of DOS Prediction in Trajectory Prediction Networks.} This study pioneers the integration of DOS prediction with advanced trajectory prediction networks, enhancing the robustness and accuracy of autonomous vehicle navigation under uncertain conditions.

    \item \textbf{Mathematical Model of DOS.} The research introduces an innovative elliptic set representation of DOS, enabling precise and flexible depiction of spatial uncertainties in traffic participant movements.
    
    \item \textbf{Optimized Objectives for DOS Prediction.} 
    The paper sets clear design objectives for DOS, emphasizing \textbf{coverage} and \textbf{area} to enhance the safety and efficiency of autonomous driving. A specialized training strategy aligns with these goals, ensuring optimal balance between safety and operational efficiency.
    
\end{itemize}

\section{Related Work}

\subsection{Trajectory Prediction Networks}

In autonomous driving, the accuracy of trajectory prediction networks is crucial for ensuring traffic safety \cite{bharilya2023machine, deng2023deep}. These networks use deep learning to process complex time-series data, predicting vehicle and pedestrian movements over various time scales. Recurrent Neural Networks (RNNs) \cite{min2024hierarchical} and Convolutional Neural Networks (CNNs) \cite{zamboni2022pedestrian} manage temporal sequences and spatial features effectively. Additionally, social and interaction models \cite{zhao2022interaction, ge2023causal} capture dynamics like pedestrian avoidance and vehicle coordination. The integration of attention mechanisms and graph neural networks \cite{lv2023ssagcn, xu2022adaptive} further enhances the accuracy and robustness in dense traffic scenarios. Despite these advances, trajectory predictions often contain errors, particularly in complex or unfamiliar environments. To mitigate this, our study introduces DOS prediction to improve reliability in vehicle planning and decision-making.

\subsection{Online Estimation for Trajectory Prediction Performance}

Recent research highlights the importance of online monitoring algorithms to address deficiencies in trajectory prediction models. These methods include uncertainty-based approaches \cite{gawlikowski2023survey, abdar2021review}, anomaly detection techniques \cite{bogdoll2022anomaly}, and error estimation methods \cite{sun2022see}, which help quantify prediction uncertainty \cite{shao2023failure}, detect potential anomalies or outliers \cite{han2022adbench}, and calculate prediction error margins \cite{shao2024when}, respectively.  These tools improve the robustness and adaptability of autonomous driving systems to unpredictable traffic conditions. 
Nonetheless, existing methods have not thoroughly explored how to compensate for deficiencies in trajectory prediction results based on estimation outcomes. To address this challenge, this study introduces a novel method for predicting DOS, providing an effective means to overcome the limitations of trajectory prediction models. This method preserves the prediction accuracy of existing trajectory prediction networks while transforming predicted trajectories into a more reliable and trustworthy form.

\subsection{Set-based Prediction}

The prediction of occupancy sets is fundamental in assessing autonomous driving safety. Traditionally, these predictions rely on physical models and formalized traffic rules \cite{koschi2020set, pek2020using, koschi2017spot}, suitable for static and predictable scenarios. With advances in technology, new approaches using probabilistic models and machine learning have emerged to predict occupancy sets in more dynamic and uncertain contexts \cite{geisslinger2021watch, geisslinger2023ethical}. These methods provide a more adaptable representation of potential occupancies but may lack precise information about the extent of these sets. Additionally, self-aware prediction model \cite{shao2023self} estimates prediction errors to monitor performance and define error bounds.
Overall, probabilistic prediction models that aim for maximum likelihood estimation (MLE) and error estimation models guided by actual prediction errors do not guarantee maximum coverage of actual future trajectories within the smallest possible space, which is critical for constructing reliable DOS.

Our research introduces a novel learning-based DOS prediction method, which is more flexible and effective than traditional methods. We also propose new evaluation metrics—coverage and occupancy set area—to provide clearer assessment criteria. A multi-objective optimization training strategy is implemented to enhance both the safety and efficiency of autonomous driving.

\section{DOS Prediction}

\subsection{Trajectory Prediction}

\begin{figure}
    \centering
    \includegraphics[width=8cm]{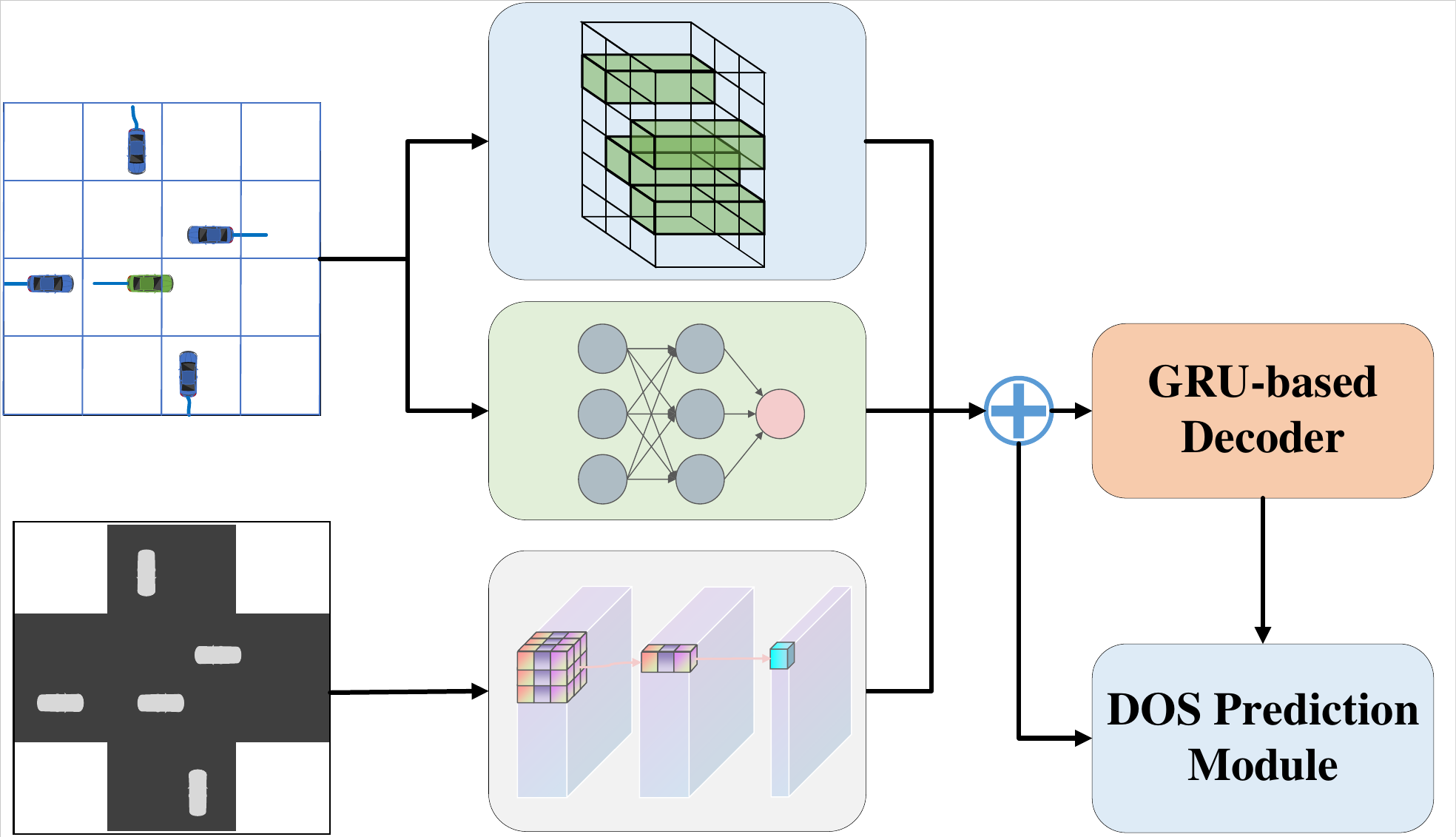}
    \caption{Trajectory-prediction-aware DOS prediction model.}
    \label{fig2}
\end{figure}


Our research builds upon the rapidly evolving field of trajectory prediction, leveraging its advanced techniques while addressing the inevitable issue of prediction errors. Consequently, we propose a trajectory-prediction-aware DOS prediction model, as shown in Fig. \ref{fig2}.

In a representative trajectory prediction network \(F\), the process begins with meticulous input data preprocessing, primarily focusing on historical location information \(\textbf{S} = {\textbf{s}_i^{-T_{h}:0}}\), including coordinates, velocity, and direction of the \(i\)-th traffic participant at time \(t\), where \(i\) ranges from 1 to \(N\). 
This preprocessing phase encompasses feature extraction, affine transformations of trajectory data, processing of local map data \(\mathcal{M}\), and masking of incomplete historical or future trajectories. It plays a critical role in structuring the input data for effective analysis. The prediction results \(\hat{\textbf{Y}}\) include the predicted trajectories \(\hat{\textbf{Y}}_i = \hat{\textbf{y}}_i^{1:T_{f}}\) for traffic participant \(A_{i}\), which are determined by \(\hat{\textbf{Y}} = F(S, \mathcal{M})\).

The architecture of the network \(F\) synergistically combines CNNs for spatial feature extraction with Gate Recurrent Unit (GRUs) networks for temporal data processing. Another significant aspect of this architecture lies in its ability to model interaction relationships. Specialized layers or mechanisms are integrated to interpret interdependencies and dynamic interactions among multiple trajectories sharing the same space. This interaction modeling is pivotal for comprehending complex spatial-temporal patterns, especially in environments with interacting entities, such as densely populated urban areas or intricate traffic systems.

The trajectory prediction model undergoes rigorous training using extensive datasets that encompass a wide spectrum of movement patterns, enabling it to learn and adapt to various behaviors. Key training techniques, including backpropagation and gradient descent, are employed to refine network weights. Hyperparameter tuning, involving adjustments to learning rates, batch sizes, and network layers, is meticulously conducted to optimize performance.

\subsection{DOS Representation} \label{sec:DOS Representation}

Our approach relies on mathematical representations, specifically ellipses, which provide a flexible and computationally feasible means of defining  DOS. These ellipses serve as a structured and interpretable form for capturing the uncertainty associated with the future positions of traffic participants. Formally, for each traffic participant \(A_{i}\), the DOS at time \(t\) is denoted as \(\mathcal{O}_i^t\), where \(\mathcal{O}_i^t\) is characterized by its center \(\mu_i^t\), major axis length \(l_i^t\), minor axis length \(w_i^t\), and orientation angle \(\theta_i^t\), with respect to the horizontal axis.

The ellipse center \(\mu_i^t\) represents the most probable position of the traffic participant at time \(t\), which is extracted from the predicted trajectory. The major and minor axis lengths, \(l_i^t\) and \(w_i^t\), determine the size of the ellipse, representing the uncertainty in both longitudinal and lateral directions. The orientation angle \(\theta_i^t\) captures the direction of this uncertainty.
Different constraints can be applied to achieve diverse variants of DOS based on varying assumptions:

\begin{itemize}
    \item Simultaneous Constraints on Axis Length and Orientation Angle: Assuming equal axis lengths (\(l_i^t=w_i^t\)) and a fixed orientation angle (\(\theta_i^t=0\)) simplifies the ellipse into a standard circle. This representation provides a conservative estimate.
    \item Constraints on Orientation Angle: Depending on different kinematic assumptions, the orientation angle \(\theta_i^t\) of the ellipse can be defined. For example, fixed angles (0 or \(\pi/4\)), angles aligned with the current velocity direction, or angles aligned with the predicted trajectory direction can be specified.
    \item Unrestricted: No constraints are imposed on the axis lengths and orientation angle. This represents the most flexible estimation form.
\end{itemize}

These DOS representations allow us to effectively capture and manage uncertainty in the context of traffic participant movements, providing a structured and interpretable framework for safety-aware autonomous systems.

\subsection{DOS Prediction Module}

\begin{figure}
    \centering
    \includegraphics[width=8cm]{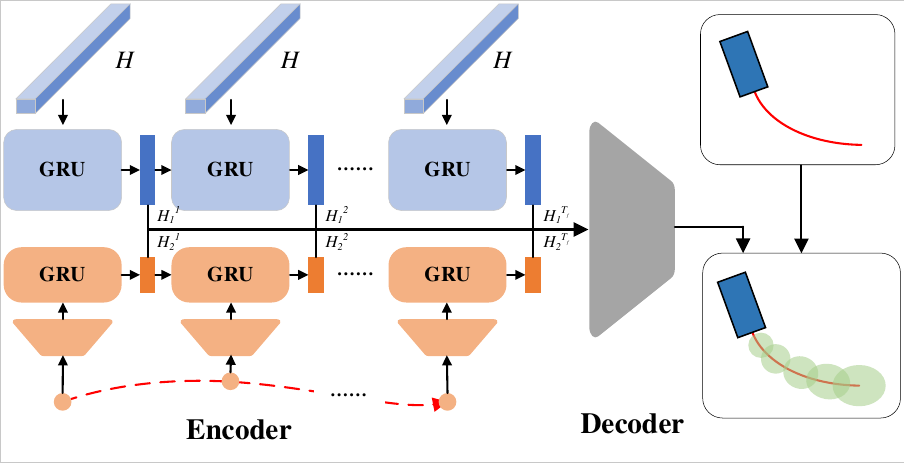}
    \caption{Design of DOS prediction module.}
    \label{fig:3}
\end{figure}

The core of our approach is centered on the development of a DOS prediction module that utilizes information from the trajectory prediction network. This module is tasked with generating the DOS for traffic participants across the prediction horizon \(T_f\), represented as 
\(\mathcal{O}_{1:N}^{1:T_{f}}.\)

Our model architecture is based on neural networks, which have demonstrated remarkable effectiveness in various domains for capturing complex relationships.
For each traffic participant, our model's input includes the output features \(\textbf{H}\) from the encoding layer of the trajectory prediction network and the predicted trajectory \(\hat{\textbf{Y}}_{i}\). \(\textbf{H}\) aggregates the encoding of all input information from the trajectory prediction network, such as modeling of maps, interactions, and historical data, thus providing crucial references for DOS prediction. \(\hat{\textbf{Y}}_{i}\) serves as supplementary information.

Given the advantages of GRUs in capturing temporal dependencies and sequence patterns, we employ GRUs as the primary structure in the DOS prediction module. As illustrated in Fig. \ref{fig:3}, \(\textbf{H}\) is first duplicated along the time dimension to serve as input to the GRU-based encoder, resulting in encoded features \(\textbf{H}_1\). On the other hand, \(Y_{i}\) goes through a linear layer and a GRU-based encoder to obtain features \(\textbf{H}_2\). Subsequently, \(\textbf{H}_1\) and \(\textbf{H}_2\) are temporally concatenated as input to the decoder for DOS prediction. Furthermore, specific activation functions are designed to ensure that the predicted ellipse parameters, including axis lengths and orientation angles, fall within reasonable ranges.

Our DOS prediction module combines the encoded information from the trajectory prediction network with the predicted trajectory details, enabling it to generate DOS that accounts for both individual behaviors and interaction patterns among traffic participants. This integration enhances the robustness and accuracy of our approach in addressing uncertainty in complex scenarios.

\subsection{Design Objectives}

In our research, the design of DOS is guided by specific objectives aimed at providing a robust representation of future traffic scenarios while prioritizing safety and efficiency. These design objectives play a pivotal role in shaping the characteristics of the generated DOS. In this section, we delineate our design objectives and elucidate their significance.

\begin{enumerate}
    \item \textbf{Sufficient Coverage:} A primary design objective is to ensure that the generated DOS offers ample coverage. This entails effectively enveloping the real future trajectories of traffic participants. Adequate coverage is imperative for furnishing a dependable estimation of prediction errors and fortifying overall driving safety and reliability. By encompassing a broad spectrum of feasible positions, DOS serves as a bulwark against uncertainties and unanticipated behaviors.
    \item \textbf{Minimal DOS Area:} Another pivotal design objective is to minimize the range of DOS. This objective is underscored by the aim to maximize the navigable space within the prediction horizon. A reduced DOS range translates to an expanded area for autonomous vehicle planning and maneuvering. By optimizing accessible space, our goal is to enhance planning feasibility and efficiency, thereby facilitating more agile and adaptive driving decisions.
\end{enumerate}

Balancing these objectives is imperative, as an overly conservative approach may result in unnecessarily large DOS, thereby impacting efficiency. Conversely, an overly aggressive approach may compromise safety. Our training and optimization strategies endeavor to strike a harmonious equilibrium that achieves both adequate coverage and a minimal DOS range, thereby fostering safe and efficient autonomous driving in dynamic environments.

To achieve the above objectives, we design specific loss functions to separately consider coverage and area:
\begin{equation}
    L = L_{c} + kL_{a},
    \label{eq1}
\end{equation}
where \(k\) represents a weighting factor used to balance the importance assigned to these two categories of objectives.

\(L_{c}\) is formulated based on the coverage of DOS. For each time step and each predicted object, we construct a multivariate gaussian distribution based on the predicted DOS, with its mean and covariance determined by the center and shape of \(\mathcal{O}_i^t\). Coverage is then assessed by calculating the Mahalanobis distance \(d_{M}\) between the true position \(\textbf{p}_{i}^{t}\) and the center of the distribution. By encouraging \(d_{M}\) to decrease, we aim to boost coverage, especially when \(d_{M} > 1\), indicating that the DOS does not envelop \(\textbf{p}_{i}^{t}\), and in such cases, it should be penalized more rigorously.
Specifically:

\begin{equation}
    {L}_{c} = \frac{1}{NT}{\sum }_{i = 1}^{N}{\sum }_{t = 1}^{T}{{l}_{c,i}^{t}},
\end{equation}

\begin{equation}
    {l}_{c,i}^{t} = \begin{cases}
        {d}_{M} & {d}_{M} \leq 1 \\
        {{d}_{M}}^\alpha + \beta & {d}_{M} > 1
    \end{cases},
    \label{eq2}
\end{equation}
where \(\alpha\) and \(\beta\) represent factors that can be manually set or obtained through data-driven statistics. 

\(L_{a}\) is obtained by calculating the area of the ellipses:
\begin{equation}
    {L}_{a} = \frac{\pi}{4} \times \frac{1}{NT}{\sum }_{i = 1}^{N}{\sum }_{t = 1}^{T}{l_{i}^{t} \times w_{i}^{t}}.
\end{equation}

Furthermore, the optimization of the DOS prediction module should not come at the expense of sacrificing the accuracy of the trajectory prediction model. Therefore, we employ parameter freezing techniques to ensure that the parameters of the trajectory prediction network remain unchanged during the optimization of this module. In practice, a two-stage training approach, involving first training the trajectory prediction network and then training the DOS prediction module, enhances the flexibility and transferability of this module's training.
Additionally, the training process utilizes an optimizer based on Adam, a step-wise learning rate adjustment strategy, and a masking strategy for insufficient trajectory lengths to optimize model performance.

\section{Experimental Setup}

\subsection{Datasets}
To evaluate our DOS prediction model, we utilize the CommonRoad dataset \cite{geisslinger2021watch}, which offers a wide variety of realistic traffic scenarios. It serves as the primary source for our evaluations, which includes urban traffic, highway scenarios, and complex intersection environments, providing a diverse range of traffic behaviors, vehicle types, and interaction patterns.
Additionally, we introduce the SIND dataset \cite{xu2022drone}, focusing specifically on signalized intersections, which are among the most challenging scenarios for autonomous driving. Located in Tianjin, China, SIND captures the complexities and stochastic nature of intersections. This dataset serves as a valuable supplement to our evaluation, particularly for assessing our model's performance in intersection scenarios.


\subsection{Evaluation Metrics}
In the evaluation of the DOS prediction framework, we employ two key metrics: \textbf{Coverage Rate (CR)} and \textbf{Occupancy Set Area (OSA)}. These metrics correspond to the two design objectives we introduced earlier and offer essential insights into the approach's efficacy across various traffic scenarios.
\begin{itemize}
    \item \textbf{CR} measures the extent to which the predicted DOSs encompass the true future trajectories of traffic participants. It is calculated as:
    \begin{equation}
        CR_{t} = CR_{i}^{t}
    \end{equation}
    \begin{equation}
        CR_{i}^{t} = \begin{cases}
            1 & \text{if }\textbf{p}_i^t \text{ is inside DOS at } t \\
            0 & \text{otherwise}
        \end{cases}
    \end{equation}
    \item \textbf{OSA} assesses the area covered by DOS at each time. It provides insights into the constrained prediction space for autonomous vehicles, emphasizing the navigable region. A smaller OSA indicates a more compact and efficient prediction space.
\end{itemize}

Furthermore, as a supplementary measure, we have incorporated the predicted DOS into the autonomous driving planning process to validate their practical advantages.  In this regard, DOS serves as a constraint during the autonomous vehicle's trajectory planning, ensuring that the planned path does not intersect with the DOS predictions of surrounding vehicles.  To assess the planning results, we conducted validation using 150 scenarios from the CommonRoad platform.  The evaluation criteria included \textbf{success rate}, \textbf{collision rate}, \textbf{out-of-bounds rate}, and \textbf{velocity}.

\begin{table*}[]
\centering
\caption{Comparison of CR and OSA for Different DOS Prediction Methods in CommonRoad.}
\begin{tabular}{m{4.0cm}<{\centering} | m{1.2cm}<{\centering} m{1.2cm}<{\centering} m{1.2cm}<{\centering} m{1.2cm}<{\centering} | m{1.2cm}<{\centering} m{1.2cm}<{\centering} m{1.2cm}<{\centering}  m{1.2cm}<{\centering}}
\toprule
\multirow{3}{*}{\textbf{}}     & \multicolumn{4}{c|}{\textbf{CR (\%)}} & \multicolumn{4}{c}{\textbf{OSA}}
\\ \cmidrule(l){2-9}

                        & 1s & 2s & 3s & 4s & 1s & 2s & 3s & 4s \\
\midrule

\textbf{PR-based ($z=0.5$)}                 & 41.7 & 35.2 & 32.7 & 25.4 & 0.179  & 1.109  & 2.667  & 5.309  \\
\textbf{PR-based ($z=1.0$)}                 & 76.2 & 72.5 & 71.6 & 62.7 & 0.715  & 4.434  & 10.666 & 21.237 \\
\textbf{PR-based ($z=1.5$)}                 & 89.9 & 87.7 & 88.3 & 82.9 & 1.608  & 9.976  & 23.999 & 47.784 \\
\textbf{PR-based ($z=2.0$)}                 & 95.1 & 93.8 & 94.5 & 91.9 & 2.859  & 17.737 & 42.664 & 84.949 \\
\midrule

\textbf{SA-based ($z=1.0$)}                 & 46.0  & 42.8 & 44.5 & 40.1 & 0.319  & 2.146  & 6.055  & 13.118 \\
\textbf{SA-based ($z=1.5$)}                 & 65.4 & 62.3 & 63.7 & 59.5 & 0.717  & 4.827  & 13.623 & 29.515 \\
\textbf{SA-based ($z=2.0$)}                 & 77.8 & 75.4 & 76.9 & 73.3 & 1.274  & 8.582  & 24.219 & 52.471 \\
\midrule

\textbf{Circle-based ($r=1.0$)}             & 91.7 & 75.3 & 69.5 & 62.8 & 3.142  & 3.142  & 3.142  & 3.142  \\
\textbf{Circle-based ($r=2.0$)}             & 98.1 & 87.3 & 80.0   & 72.1 & 12.566 & 12.566 & 12.566 & 12.566 \\
\textbf{Circle-v1-based ($z=0.7$)}         & 85.8 & 81.1 & 80.8 & 76.9 & 1.539  & 6.158  & 13.854 & 24.63  \\
\textbf{Circle-v1-based ($z=1.0$)}          & 91.7 & 87.3 & 86.5 & 82.6 & 3.142  & 12.566 & 28.274 & 50.265 \\
\textbf{Circle-v2-based ($z=1.0$)}          & 70.8 & 71.9 & 73.1 & 73.0  & 0.350   & 2.107  & 5.141  & 14.234 \\
\textbf{Circle-v2-based ($z=1.5$)}          & 79.2 & 78.8 & 79.3 & 78.8 & 0.787  & 4.740   & 11.567 & 32.026 \\
\midrule

\rowcolor{gray!20}
\textbf{DOS (circle)}                       & 96.0  & 88.2 & 85.3 & 78.7 & 3.369  & 7.815  & 12.633 & 18.849 \\
\rowcolor{gray!20}
\textbf{DOS (elipse, $\theta=0$)}              & 96.6 & 91.2 & 88.0  & 81.0  & 2.303  & 5.676  & 9.004  & 13.087 \\
\rowcolor{gray!20}
\textbf{DOS (elipse, $\theta=\pi/4$)}             & 95.8 & 87.7 & 84.6 & 78.1 & 3.442  & 7.591  & 12.033 & 17.844 \\
\rowcolor{gray!20}
\textbf{DOS (elipse, v1)}     & 96.7 & 90.4 & 87.0  & 80.0   & 2.869  & 6.191  & 9.483  & 13.813 \\
\rowcolor{gray!20}
\textbf{DOS (elipse, v2)}   & 96.8 & 90.9 & 87.7 & 80.7 & 2.453  & 5.706  & 9.088  & 13.448 \\
\rowcolor{gray!20}
\textbf{DOS (elipse)}                       & 96.9 & 91.2 & 87.8 & 81.1 & 2.461  & 5.827  & 9.147  & 13.364 \\

\bottomrule 
\end{tabular}
\label{Table1}
\end{table*}

\begin{table}[]
\centering
\caption{Comparison of Different Methods in SIND.}

\begin{tabular}{m{3.4cm}<{\centering} | m{0.7cm}<{\centering} m{0.7cm}<{\centering} | m{0.7cm}<{\centering} m{0.7cm}<{\centering}}
\toprule
\multirow{3}{*}{\textbf{}}     & \multicolumn{2}{c|}{\textbf{CR(\%)}} & \multicolumn{2}{c}{\textbf{OSA}}
\\ \cmidrule(l){2-5}

                        & 1s & 4s & 1s & 4s \\
\midrule

\textbf{PR-based ($z=2.0$)}                 & 63.8  & 45.6 & 1.841  & 38.55  \\

\textbf{SA-based ($z=1.0$)}                 & 94.1 & 77.1 & 1.956 & 11.469 \\

\textbf{Circle-based ($r=1.0$)}             & 93.7  & 60.4 & 3.142  & 3.142  \\

\textbf{Circle-v1-based ($z=1.0$)}          & 93.7  & 86.9 & 3.142  & 50.265 \\

\textbf{Circle-v2-based ($z=1.5$)}          & 83.0  & 79.3 & 0.872  & 21.051 \\
\midrule

\rowcolor{gray!20}
\textbf{DOS (elipse, $\theta=0$)}  & 95.8 & 79.1  & 1.590    & 9.885 \\
\rowcolor{gray!20}
\textbf{DOS (elipse, $\theta=\pi/4$)} & 94.2 & 77.8  & 2.106  & 11.542 \\
\rowcolor{gray!20}
\textbf{DOS (elipse, v1)}                   & 94.0  & 77.9 & 2.094  & 11.088 \\
\rowcolor{gray!20}
\textbf{DOS (elipse, v2)}                   & 95.3  & 79.4 & 1.874  & 10.421 \\
\rowcolor{gray!20}
\textbf{DOS (elipse)}                       & 95.6  & 78.6 & 1.617  & 9.751 \\

\bottomrule 
\end{tabular}
\label{Table2}
\end{table}

\subsection{Comparative Methods}

We compare the proposed DOS prediction framework against several comparative methods, each offering a distinct approach to estimating DOS:

\begin{itemize}
    \item \textbf{Probability Prediction (PR)-Based Method\cite{geisslinger2021watch}:}
    This method models the possible positions of traffic participants at each future time step as Gaussian distributions. These probabilistic models are trained using MLE. To convert these probability distributions into DOSs with well-defined boundaries, the method forms ellipsoidal occupancy sets centered around the mean of the Gaussian distribution. The standard deviation of the Gaussian distribution is scaled by a factor \(z\) to estimate DOSs with different ranges. Specifically, we consider three values of \(z\): 0.5, 1, and 2.

    \item \textbf{Self-Aware Trajectory Prediction (SA)-Based Method \cite{shao2023self}:}
    This method adopts a training strategy guided by fitting the predicted errors to estimate occupancy sets. In this approach, the predicted error is used as the radius of a circular occupancy set. Similarly, a scaling factor \(z\) is introduced to scale the circular occupancy set.

    \item \textbf{Specified Occupancy Set:}
    In this method, we assume that occupancy sets are circular. The radius for each future time step is manually specified. Three variants are considered: 1) Circle-based, where a fixed radius \(r\) is set for each time step; 2) Linear-based, where the radius increases linearly with time, i.e., \(r=z \times t\), where \(z\) is a scaling factor; 3) Error-based, where the initial radius \(r_{e}\) for each time step is set based on the average error statistics, and the occupancy set radius is then \(r=z \times r_{e}\).

\end{itemize}

For our proposed methods, we also implemented different variants based on the representations proposed in Section \ref{sec:DOS Representation}: 1) DOS (circle), assuming DOS as a circle; 2) DOS (ellipse, \(\theta=\theta_0\)), assuming DOS as an ellipse with a fixed orientation after affine transformations; 3) DOS (ellipse, v1), assuming the ellipse's orientation is determined by the current velocity direction; 4) DOS (ellipse, v2), assuming the ellipse's orientation is determined by the predicted velocity direction at each time. 5) DOS (ellipse), where the ellipse's orientation is also estimated by the DOS prediction module.

Additionally, for comparative purposes, we compute the optimal DOS assuming circular shapes or ellipses with specified orientation angles. These optimal occupancy sets are determined by constraining the Mahalanobis distance of \(\textbf{p}_{i}^{t}\) to the ellipses to be equal to 1.

\section{Results and Discussion}

\begin{figure*}
    \centering
    \begin{subfigure}{3.4cm}
        \centering
        \includegraphics[width=\textwidth]{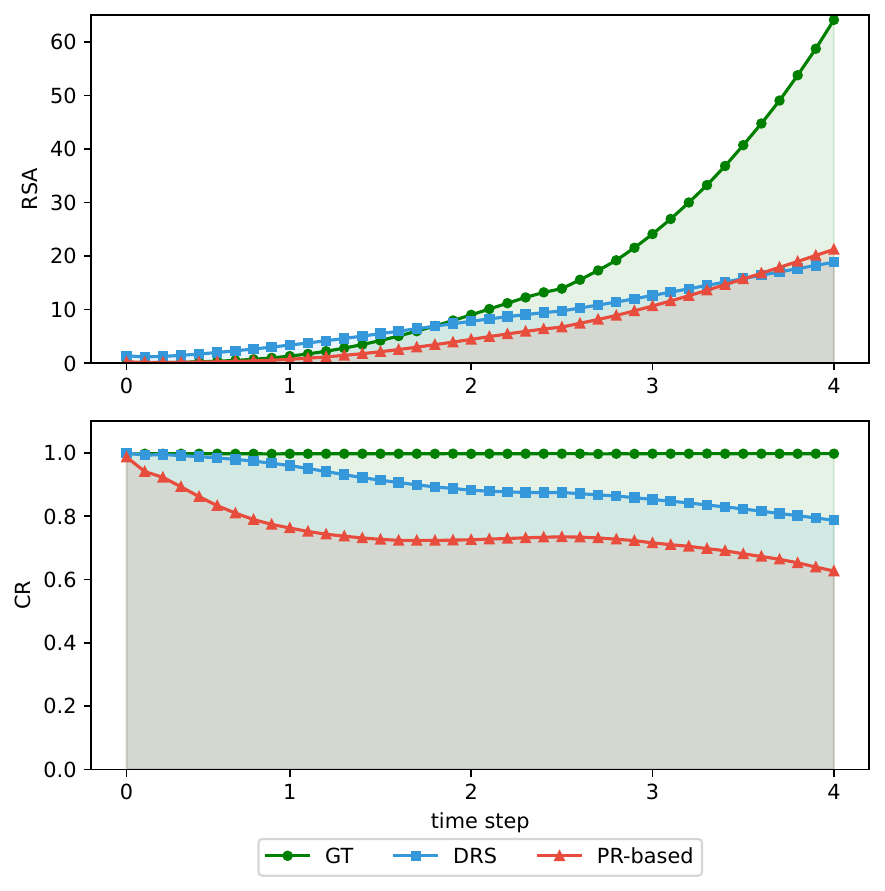}
        \caption{Circle}
    \end{subfigure}
    \begin{subfigure}{3.4cm}
        \centering
        \includegraphics[width=\textwidth]{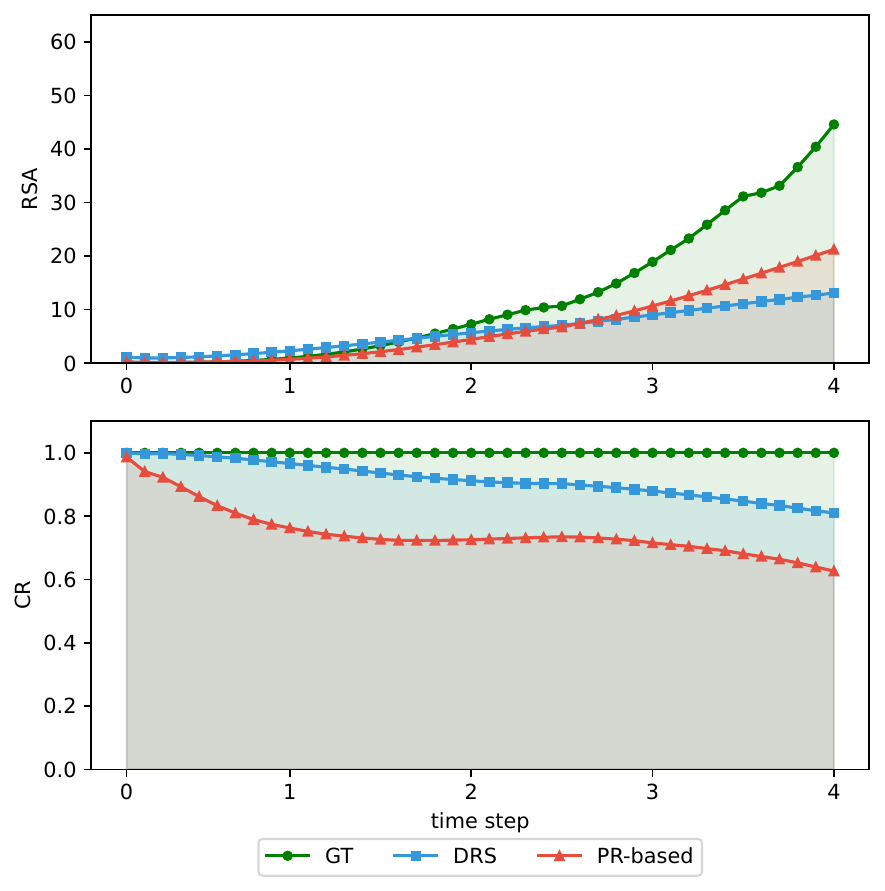}
        \caption{Elipse, $\theta=0$}
    \end{subfigure}
    \begin{subfigure}{3.4cm}
        \centering
        \includegraphics[width=\textwidth]{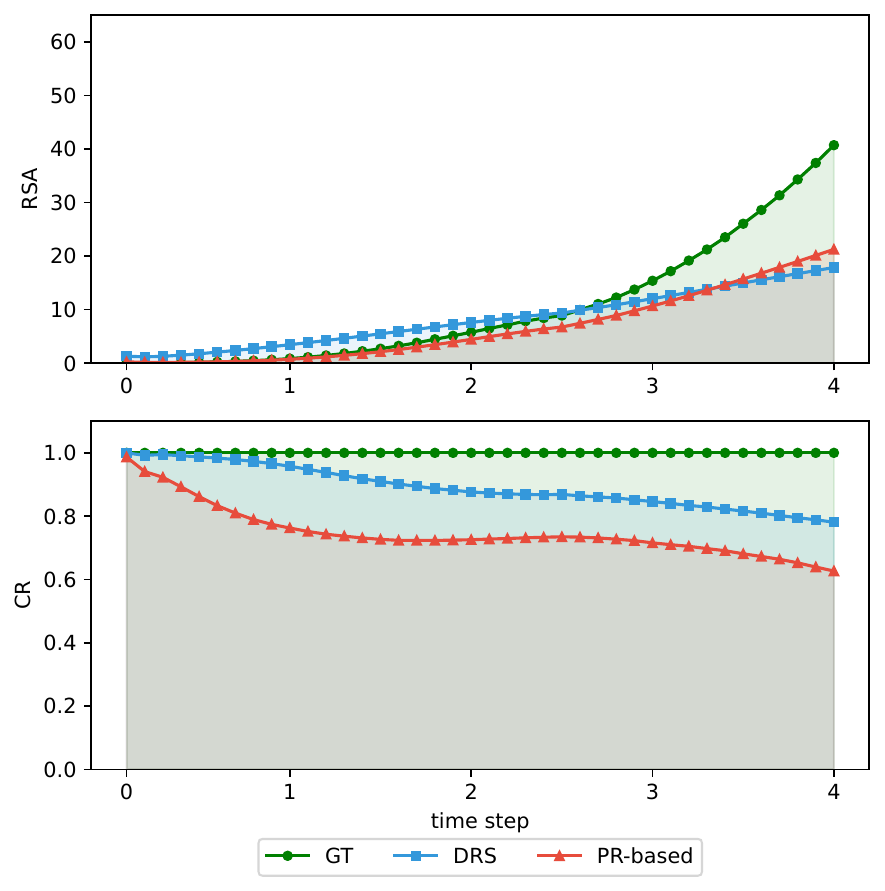}
        \caption{Elipse, $\theta=\pi/4$}
    \end{subfigure}
    \begin{subfigure}{3.4cm}
        \centering
        \includegraphics[width=\textwidth]{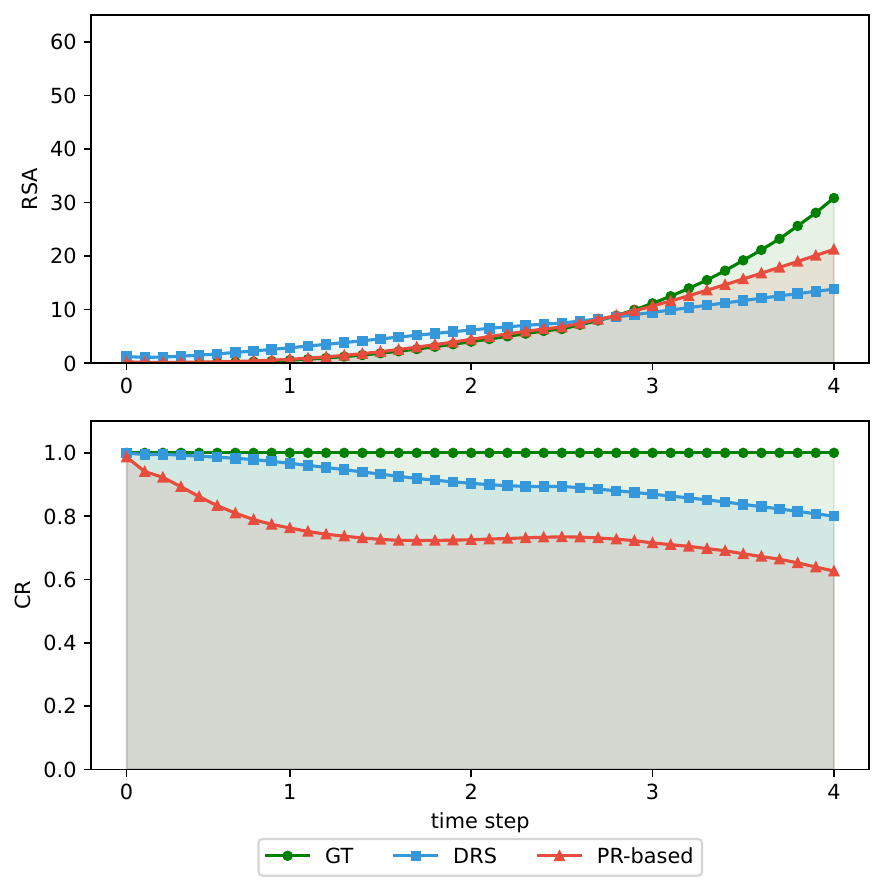}
        \caption{Elipse, v1}
    \end{subfigure}
    \begin{subfigure}{3.4cm}
        \centering
        \includegraphics[width=\textwidth]{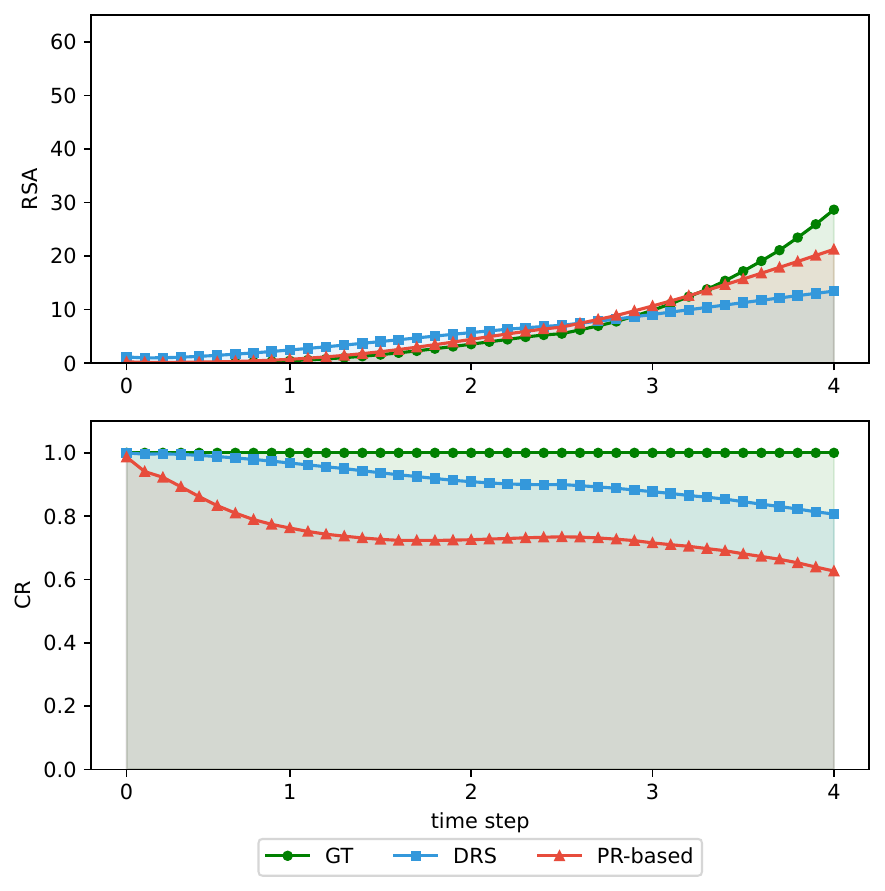}
        \caption{Elipse, v2}
    \end{subfigure}
        
    \caption{Analysis Different DOS Representations.}
    \label{figr1}
\end{figure*}

    
        

\subsection{Comparing Different Methods}

In this section, we conduct a comprehensive comparison of various methods for estimating DOS and assess their performance using two pivotal metrics: CR and OSA across different prediction horizons, as in Table \ref{Table1}. Our goal is to gain insights into how each method balances safety and efficiency considerations in autonomous driving scenarios. Here are the key findings:

The PR-based method, commonly used in trajectory prediction, exhibits consistent but moderate CR and OSA values for varying scaling factors (\(z\)). This approach provides practitioners with flexibility to fine-tune the scaling factor to meet specific requirements. Notably, an increase in \(z\) results in higher CR and OSA, indicating that achieving a higher coverage rate comes at the cost of a more conservative estimation, potentially sacrificing efficiency in certain situations.

The SA-based method, while excelling in predicting errors, may not be the most suitable for effectively balancing CR and OSA. It tends to prioritize the accuracy of predicted errors, aiming for occupancy set boundaries that closely match the average of all errors. This tendency results in nearly half of the coverage rate at \(z=1\). This further highlights the necessity of our study, which focuses on achieving the objectives of sufficient coverage and Minimal DOS Area.

The specified occupancy set Method, based on predefined assumptions, serves as a rule-based baseline. It demonstrates reasonable performance when appropriate parameters are chosen. However, achieving higher coverage rates may require larger occupancy set area. Fine-tuning of parameters may also be necessary to strike the desired balance.

Our proposed DOS methods, encompassing circular and elliptical shapes with various orientation representations, consistently outperform other methods, offering significantly higher CR while maintaining smaller OSA values. These results underscore the efficacy of our approach in providing comprehensive coverage while minimizing the area of the occupancy sets—a critical factor for ensuring both safety and efficiency in autonomous driving scenarios.
In comparison to the assumptions of circular DOS, the elliptical assumptions consistently deliver superior results.  Furthermore, although certain methods, such as DOS (ellipse, \(\theta=0)\), demonstrate promising performance, considering the complexity of real-world scenarios and the flexibility of our approach, DOS (ellipse) exhibits broader applicability.

Furthermore, we assessed the performance of our DOS methods on the SIND dataset, representing complex real-world traffic scenarios. Due to space constraints, we presented only the most representative variant for each category of methods compared.
The results, presented in Table \ref{Table2}, reinforce our earlier observations, highlighting the adaptability and robustness of our methods across diverse scenarios.
Our DOS methods consistently outperformed other approaches in the SIND dataset, demonstrating their effectiveness in challenging and dynamic traffic environments. This reaffirms their potential to ensure both safety and efficiency in autonomous driving applications.

\subsection{Upper Limit Analysis of Different DOS Representation}

In this subsection, we focus on an upper limit analysis of various DOS representations, with a primary objective of evaluating their performance under the condition of ensuring 100\% coverage while minimizing DOS area. To achieve this goal, the true coordinates at each time precisely align with the DOS boundaries.
For circular DOS assumptions, we determine the upper limit by setting the radius equal to the true prediction error. 
For elliptical DOS with specified orientation angles, we compute the upper limit by constraining the Mahalanobis distance to equal 1, thereby deriving the theoretical minimum OSA for these representations.
The results are shown in Fig. \ref{figr1}, which elucidates several insights:

1) Different DOS representations exhibit distinct upper limits. Circular DOS assumptions, in theory, require the largest OSA to achieve complete coverage. In contrast, fixed-orientation elliptical DOS representations offer more favorable upper limit conditions, implying a relatively smaller required OSA. Furthermore, the introduction of orientation estimation based on velocity direction further optimizes the upper limits. Additionally, for unconstrained ellipses, theoretically achieving an optimal OSA of 0 is possible.

2) The upper limit analysis prompts a reconsideration of the difficulty and necessity of achieving complete coverage. For long-term predictions, the presence of large prediction errors at certain outlier cases may result in significantly larger required OSA, potentially compromising navigable space. Therefore, under the guidance of our optimization objective, the preference leans towards predicting DOS with relatively smaller areas while ensuring adequate coverage.

3) This analysis also provides valuable insights into future directions for this research. Firstly, by optimizing the underlying trajectory prediction algorithms, we can directly improve the theoretical upper limits. Smaller trajectory prediction errors theoretically enable achieving sufficient CR with smaller OSA. Additionally, further improvements in the practical performance of DOS (ellipse, v1), DOS (ellipse, v2), and DOS (ellipse) methods are necessary to fully exploit their potential. Furthermore, Addressing long-term DOS prediction challenges requires ongoing optimization efforts.

4) Fig. \ref{figr1} includes an evaluation of the PR-based method with \(z=1\), providing a clear visual comparison that highlights the advantages of our proposed methods.

This upper limit analysis not only contributes to a deeper understanding of the capabilities and constraints of various DOS representations but also offers valuable insights into potential avenues for future research in this domain.

\subsection{Impact of Different Parameters on DOS Results}






\begin{table}[]
\centering
\caption{Results for Different Parameters.}

\begin{tabular}{m{0.8cm}<{\centering} m{0.8cm}<{\centering} | m{0.9cm}<{\centering} m{0.9cm}<{\centering} | m{0.9cm}<{\centering} m{0.9cm}<{\centering}}
\toprule
 \multirow{3}{*}{\textbf{$k$}} & \multirow{3}{*}{\textbf{$\alpha$}}     & \multicolumn{2}{c|}{\textbf{CR (\%)}} & \multicolumn{2}{c}{\textbf{OSA}}
\\ \cmidrule(l){3-6}

                &   & 1s & 4s & 1s & 4s \\
\midrule

0.5  & 3.0   & 92.1 & 71.8 & 1.049  & 6.139  \\
0.2  & 3.0   & 94.9 & 76.6 & 1.653  & 9.278  \\
0.05 & 3.0   & 98.1 & 84.3 & 3.559  & 18.047 \\
0.01 & 3.0   & 99.4 & 91.6 & 7.304  & 37.01  \\
0.01 & 10.0  & 99.8 & 96.2 & 16.706 & 96.492 \\
0.1  & 1.0   & 94.5 & 70.8 & 1.352  & 3.800  \\
0.1  & 2.0   & 95.7 & 76.7 & 1.673  & 8.093  \\
0.1  & 3.0   & 96.9 & 81.1 & 2.461  & 13.364 \\

\bottomrule 
\end{tabular}
\label{tab_ab1}
\end{table}

We conducted experiments to assess the impact of different parameter settings on the performance of our DOS prediction framework. Table \ref{tab_ab1} provides a summary of the results obtained for various combinations of the key parameters \(k\) and \(\alpha\). These experiments yield valuable insights into how these parameters influence the trade-off between coverage and efficiency:
\textbf{1) Impact of \(k\):} Our experiments revealed that smaller \(k\) values, ranging from 0.5 to 0.01, led to a significant improvement in Coverage Rate (CR), with \(k = 0.01\) achieving a remarkable 99.4\% CR at the 1-second prediction horizon. This reduction in \(k\) enhances coverage; however, it tends to result in larger OSA, affecting the compactness of DOS predictions.
\textbf{2)Impact of \(\alpha\):} Increasing the \(\alpha\) values from 1.0 to 3.0 had a positive effect on CR but came at the cost of increased OSA.

In summary, our experiments underscore the significant influence of parameter settings on the performance of our DOS prediction framework. These findings offer practitioners the flexibility to fine-tune parameters according to specific autonomous driving requirements and trade-offs.

\subsection{Planning with DOS}

Table \ref{Tab_planning} presents the outcomes of incorporating DOS into autonomous driving planning. Compared to using collision constraints solely based on predicted trajectories, the inclusion of collision constraints derived from DOS significantly enhances the safety of autonomous driving. Furthermore, DOS, with its controllable area, contributes to improved efficiency in planning. In conclusion, the introduction of DOS provides a more reliable means to ensure the safety and efficiency of autonomous driving systems.

\begin{table}[]
\centering
\caption{Results of Planning}
\begin{tabular}{m{1.2cm}<{\centering} | m{1.2cm}<{\centering} m{1.2cm}<{\centering} m{1.9cm}<{\centering} m{1.0cm}<{\centering}}

\toprule
 & \textbf{success rate (\%)} & \textbf{collision rate (\%)} & \textbf{out-of-bounds rate (\%)} & \textbf{velocity}\\
\midrule

\textit{w/o} DOS & 90.0 & 10.0 & 0.0 & 5.474 \\
\rowcolor{gray!20}
\textit{w} DOS & 94.7 & 5.3 & 0.0 & 5.710 \\

\bottomrule
\end{tabular}
\label{Tab_planning}
\end{table}

\begin{figure*}
    \centering
    \includegraphics[width=18cm]{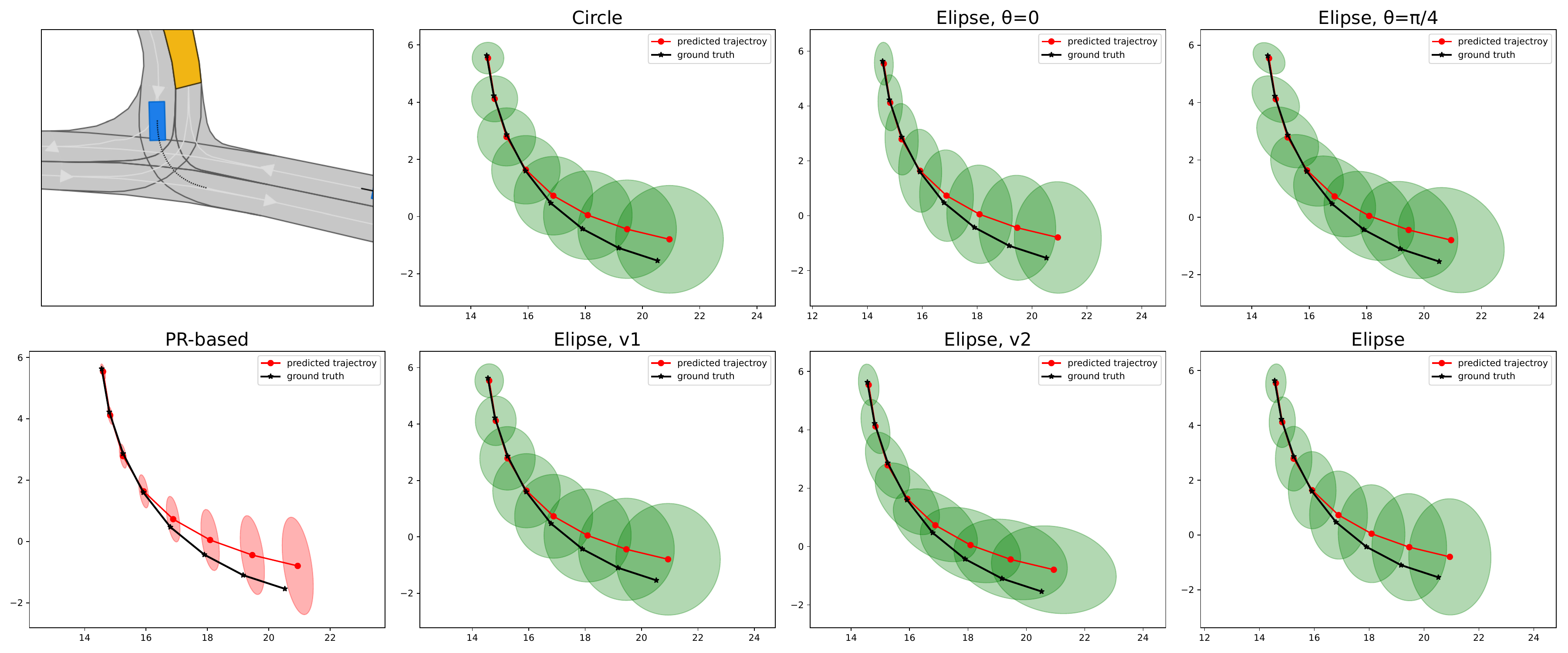}
    \caption{Visualization of Different DOS Representations. Red or green Regions Represent Predicted DOS at Each Time.}
    \label{figr2}
\end{figure*}

\subsection{Qualitative Evaluation}

To facilitate the qualitative evaluation, we provide a visual comparison in Fig. \ref{figr2}\footnote{For more results, refer to see \href{https://swb19.github.io/DOS-Prediction/}{https://swb19.github.io/DOS-Prediction/}}. This figure illustrates the performance of the proposed DOS representations compared to the PR-based method in capturing the real trajectories of a traffic participant.
The visual comparison offers an intuitive insight into how effectively each method captures the possible positions of traffic participants within the prediction horizon. In particular, it becomes evident that the PR-based method struggles to provide adequate coverage for long-term predictions, whereas the proposed DOS achieves complete coverage. Furthermore, transitioning from circular to elliptical DOS representations helps to further reduce the OSA.

\section{Conclusion}

In conclusion, this study has presented a novel method for DOS prediction. By seamlessly integrating state-of-the-art trajectory prediction networks with a purpose-built DOS prediction module, we successfully bridge the gap between predicted and actual trajectories, particularly in complex driving scenarios. Our innovative contributions encompass the development of a DOS prediction model, distinct DOS representations, and specialized evaluation metrics, all of which contribute to significantly improved safety and efficiency in autonomous driving systems. The comprehensive experimental validation conducted in this study reaffirms the superiority of our approach, providing a solid foundation for future advancements in this field.

Looking ahead, future research directions should explore expanding the state space to consider factors such as the shapes of traffic participants, offering potential refinements in DOS predictions. Additionally, the integration of DOS prediction into the autonomous driving stack to enhance safety in practical applications remains a critical avenue for further exploration.

\bibliographystyle{IEEEtran}
\bibliography{ref}
\vspace{-1cm}

\end{document}